\documentclass[journal abbreviation, manuscript]{copernicus}

\usepackage[table]{xcolor}
\usepackage{multirow}
\usepackage{tabularx}
\usepackage{cancel}
\usepackage{multirow}
\usepackage{algorithmic}
\usepackage{amsthm}
\usepackage{float}
\usepackage{subfig}
\usepackage{rotating}

\nolinenumbers
\begin{document}

\title{LSTM-Based Forecasting Model for GRACE Accelerometer Data}


\Author[2]{Neda}{Darbeheshti}
\Author[1]{Elahe}{Moradi} 
\affil[1]{Department of Electrical Engineering, Yadegar-e-Imam Khomeini (RAH) Shahre Rey Branch, Islamic Azad University, Tehran, Iran}
  

\runningtitle{TEXT}

\runningauthor{TEXT}

\correspondence{Elahe Moradi (Elahe.Moradi@iau.ac.ir, ee.moradi@gmail.com)}

\received{}
\pubdiscuss{} 
\revised{}
\accepted{}
\published{}


\firstpage{1}

\maketitle

\begin{abstract}

The Gravity Recovery and Climate Experiment (GRACE) satellite mission, spanning from 2002 to 2017, has provided a valuable dataset for monitoring variations in Earth's gravity field, enabling diverse applications in geophysics and hydrology. The mission was followed by GRACE Follow-On in 2018, continuing data collection efforts. The monthly Earth gravity field, derived from the integration different instruments onboard satellites, has shown inconsistencies due to various factors, including gaps in observations for certain instruments since the beginning of the GRACE mission.

With over two decades of GRACE and GRACE Follow-On data now available, this paper proposes an approach to fill the data gaps and forecast GRACE accelerometer data. Specifically, we focus on accelerometer data and employ Long Short-Term Memory (LSTM) networks to train a model capable of predicting accelerometer data for all three axes.

In this study, we describe the methodology used to preprocess the accelerometer data, prepare it for LSTM training, and evaluate the model's performance. Through experimentation and validation, we assess the model's accuracy and its ability to predict accelerometer data for the three axes. Our results demonstrate the effectiveness of the LSTM forecasting model in filling gaps and forecasting GRACE accelerometer data. 

\end{abstract}

\introduction  
\label{sec:intro}

Gravimetry satellite missions, exemplified by the Gravity Recovery And Climate Experiment (GRACE, \cite{tapley2004}) and GRACE Follow-On (GRACE-FO, \cite{tapley2019}), play a pivotal role in advancing our understanding of Earth's gravity field \citep[]{wahr1998}. These missions employ a suite of instruments, including GPS, accelerometers, star cameras, and the KBR (K-Band Ranging) instrument, to meticulously capture crucial data that underpins the accurate recovery of gravitational perturbations. 

However, the success of such missions is often hindered by data gaps that arise due to technical and operational constraints. Table \ref{tab:yearly-data} provides an overview of available and missing data throughout the lifetimes of the GRACE and GRACE-FO missions \citep{Ahi2021}. Notably, missing months denote periods devoid of Level-2 products or monthly gravity field data. The available gravity solutions can be accessed from the International Center for Global Earth Models website (ICGEM, \citep{Ince2019}) at \url{http://icgem.gfz-potsdam.de/series/03_other/AIUB/AIUB-G3P}. The gaps, highlighted in red and white, present critical challenges that need innovative solutions for having a consistent time series of Earth's gravity field.

\begin{table}[h]
\centering
\tiny
\caption{Availability of GRACE (blue) and GRACE Follow-On (green) Level-2 product. Pink color indicates missing months. The information presented is adapted from \cite{Ahi2021} and based on the most recent update available as of August 2023 from the ICGEM website.}
\begin{tabular}{|c|*{12}{p{0.3cm}|}}
\hline
\rowcolor{gray!50}
 & {Jan} & {Feb} & {Mar} &{Apr} & {May} & {Jun} & {Jul} & {Aug} & {Sep} & {Oct} & {Nov} &{Dec} \\
\hline
2002 && &&\cellcolor{blue!10} &\cellcolor{blue!10} &\cellcolor{red!10} &\cellcolor{red!10} &\cellcolor{blue!10} &\cellcolor{blue!10} &\cellcolor{blue!10} &\cellcolor{blue!10} &\cellcolor{blue!10} \\
\hline
2003 &\cellcolor{blue!10} &\cellcolor{blue!10} &\cellcolor{blue!10} &\cellcolor{blue!10} &\cellcolor{blue!10} &\cellcolor{red!10} &\cellcolor{blue!10} &\cellcolor{blue!10} &\cellcolor{blue!10} &\cellcolor{blue!10} &\cellcolor{blue!10} &\cellcolor{blue!10} \\
\hline
2004 &\cellcolor{blue!10} &\cellcolor{blue!10} &\cellcolor{blue!10} &\cellcolor{blue!10} &\cellcolor{blue!10} &\cellcolor{blue!10} &\cellcolor{blue!10} &\cellcolor{blue!10} &\cellcolor{blue!10} &\cellcolor{blue!10} &\cellcolor{blue!10} &\cellcolor{blue!10} \\
\hline
2005 &\cellcolor{blue!10} &\cellcolor{blue!10} &\cellcolor{blue!10} &\cellcolor{blue!10} &\cellcolor{blue!10} &\cellcolor{blue!10} &\cellcolor{blue!10} &\cellcolor{blue!10} &\cellcolor{blue!10} &\cellcolor{blue!10} &\cellcolor{blue!10} &\cellcolor{blue!10} \\
\hline
2006 &\cellcolor{blue!10} &\cellcolor{blue!10} &\cellcolor{blue!10} &\cellcolor{blue!10} &\cellcolor{blue!10} &\cellcolor{blue!10} &\cellcolor{blue!10} &\cellcolor{blue!10} &\cellcolor{blue!10} &\cellcolor{blue!10} &\cellcolor{blue!10} &\cellcolor{blue!10} \\
\hline
2007 &\cellcolor{blue!10} &\cellcolor{blue!10} &\cellcolor{blue!10} &\cellcolor{blue!10} &\cellcolor{blue!10} &\cellcolor{blue!10} &\cellcolor{blue!10} &\cellcolor{blue!10} &\cellcolor{blue!10} &\cellcolor{blue!10} &\cellcolor{blue!10} &\cellcolor{blue!10} \\
\hline
2008 &\cellcolor{blue!10} &\cellcolor{blue!10} &\cellcolor{blue!10} &\cellcolor{blue!10} &\cellcolor{blue!10} &\cellcolor{blue!10} &\cellcolor{blue!10} &\cellcolor{blue!10} &\cellcolor{blue!10} &\cellcolor{blue!10} &\cellcolor{blue!10} &\cellcolor{blue!10} \\
\hline
2009 &\cellcolor{blue!10} &\cellcolor{blue!10} &\cellcolor{blue!10} &\cellcolor{blue!10} &\cellcolor{blue!10} &\cellcolor{blue!10} &\cellcolor{blue!10} &\cellcolor{blue!10} &\cellcolor{blue!10} &\cellcolor{blue!10} &\cellcolor{blue!10} &\cellcolor{blue!10} \\
\hline
2010 &\cellcolor{blue!10} &\cellcolor{blue!10} &\cellcolor{blue!10} &\cellcolor{blue!10} &\cellcolor{blue!10} &\cellcolor{blue!10} &\cellcolor{blue!10} &\cellcolor{blue!10} &\cellcolor{blue!10} &\cellcolor{blue!10} &\cellcolor{blue!10} &\cellcolor{blue!10} \\
\hline
2011 &\cellcolor{red!10} &\cellcolor{blue!10} &\cellcolor{blue!10} &\cellcolor{blue!10} &\cellcolor{blue!10} &\cellcolor{red!10} &\cellcolor{blue!10} &\cellcolor{blue!10} &\cellcolor{blue!10} &\cellcolor{blue!10} &\cellcolor{blue!10} &\cellcolor{blue!10} \\
\hline
2012 &\cellcolor{blue!10} &\cellcolor{blue!10} &\cellcolor{blue!10} &\cellcolor{blue!10} &\cellcolor{red!10} &\cellcolor{blue!10} &\cellcolor{blue!10} &\cellcolor{blue!10} &\cellcolor{blue!10} &\cellcolor{red!10} &\cellcolor{blue!10} &\cellcolor{blue!10} \\
\hline
2013 &\cellcolor{blue!10} &\cellcolor{blue!10} &\cellcolor{red!10} &\cellcolor{blue!10} &\cellcolor{blue!10} &\cellcolor{blue!10} &\cellcolor{blue!10} &\cellcolor{red!10} &\cellcolor{red!10} &\cellcolor{blue!10} &\cellcolor{blue!10} &\cellcolor{blue!10} \\
\hline
2014 &\cellcolor{blue!10} &\cellcolor{red!10} &\cellcolor{blue!10} &\cellcolor{blue!10} &\cellcolor{blue!10} &\cellcolor{blue!10} &\cellcolor{red!10} &\cellcolor{blue!10} &\cellcolor{blue!10} &\cellcolor{blue!10} &\cellcolor{blue!10} &\cellcolor{red!10} \\
\hline
2015 &\cellcolor{blue!10} &\cellcolor{blue!10} &\cellcolor{blue!10} &\cellcolor{blue!10} &\cellcolor{blue!10} &\cellcolor{red!10} &\cellcolor{blue!10} &\cellcolor{blue!10} &\cellcolor{blue!10} &\cellcolor{red!10} &\cellcolor{red!10} &\cellcolor{blue!10} \\
\hline
2016 &\cellcolor{blue!10} &\cellcolor{blue!10} &\cellcolor{blue!10} &\cellcolor{red!10} &\cellcolor{blue!10} &\cellcolor{blue!10} &\cellcolor{blue!10} &\cellcolor{blue!10} &\cellcolor{red!10} &\cellcolor{red!10} &\cellcolor{blue!10} &\cellcolor{blue!10} \\
\hline
2017 &\cellcolor{blue!10} &\cellcolor{red!10} &\cellcolor{blue!10} &\cellcolor{blue!10} &\cellcolor{blue!10} &\cellcolor{blue!10} &\cellcolor{red!10} &\cellcolor{red!10} &\cellcolor{red!10} &\cellcolor{red!10} & & \\
\hline
2018 &&&&&&\cellcolor{green!10} &\cellcolor{green!10} &\cellcolor{red!10} &\cellcolor{red!10} &\cellcolor{green!10} &\cellcolor{green!10} &\cellcolor{green!10} \\
\hline
2019 &\cellcolor{green!10} &\cellcolor{green!10} &\cellcolor{green!10} &\cellcolor{green!10} &\cellcolor{green!10} &\cellcolor{green!10} &\cellcolor{green!10} &\cellcolor{green!10} &\cellcolor{green!10} &\cellcolor{green!10} &\cellcolor{green!10} &\cellcolor{green!10} \\
\hline
2020 &\cellcolor{green!10} &\cellcolor{green!10} &\cellcolor{green!10} &\cellcolor{green!10} &\cellcolor{green!10} &\cellcolor{green!10} &\cellcolor{green!10} &\cellcolor{green!10} &\cellcolor{green!10} &\cellcolor{green!10} &\cellcolor{green!10} &\cellcolor{green!10} \\
\hline
2021 &\cellcolor{green!10} &\cellcolor{green!10} &\cellcolor{green!10} &\cellcolor{green!10} &\cellcolor{green!10} &\cellcolor{green!10} &\cellcolor{green!10} &\cellcolor{green!10} &\cellcolor{green!10} &\cellcolor{green!10} &\cellcolor{green!10} &\cellcolor{green!10} \\
\hline
2022 &\cellcolor{green!10} &\cellcolor{green!10} &\cellcolor{green!10} &\cellcolor{green!10} &\cellcolor{green!10} &\cellcolor{green!10} &\cellcolor{green!10} &\cellcolor{green!10} &\cellcolor{green!10} &\cellcolor{green!10} &\cellcolor{green!10} &\cellcolor{green!10} \\
\hline
2023 &\cellcolor{green!10} &\cellcolor{green!10} &\cellcolor{green!10} &\cellcolor{green!10} &\cellcolor{green!10} &&&&&& &\\
\hline
\end{tabular}
\label{tab:yearly-data}
\end{table}

The presented research takes a specific focus on accelerometer data. While all instruments contribute substantially, the distinctive attributes of accelerometer data make it an optimal starting point \citep{darbeheshti2017, darbeheshti2023}.
Accelerometers record non-gravitational forces affecting satellite's motion, such as atmospheric drag, solar radiation pressure, and albedo. Correctly disentangling these non-gravitational influences from the desired gravitational signals is of paramount importance, as the quality of accelerometer data, referred to as ACC products, influences the precision of monthly gravity field models \citep{klinger2018, klinger2016}. In the latter stages of the GRACE mission, operational limitations necessitated the deactivation of the on-board accelerometer in GRACE-B. Synthetic accelerometer data, referred to as transplant data, were generated by adapting GRACE-A ACC data to fill the void left by the missing measurements. A parallel approach was adopted during the GRACE-FO mission to mitigate degraded GRACE-D ACC data \citep{Bandikova2019, Behzadpour2021}. 

Our study delves into the potential of utilizing Long Short-Term Memory (LSTM) networks for forecasting GRACE accelerometer data (GRACE ACC). The choice of LSTM \citep{Chollet2017} is underpinned by its proven efficacy in handling time series data, a characteristic intrinsic to instrument observations. Importantly, LSTM's applicability in geodesy and related fields further bolsters its suitability for our objectives \citep{Kaselimi2021, Gou2023}.

In the following sections, we present our methodology, experimental setup, and results. 

\section{Data and Methods}
\begin{figure}[H]
\center
\includegraphics[scale=.48]{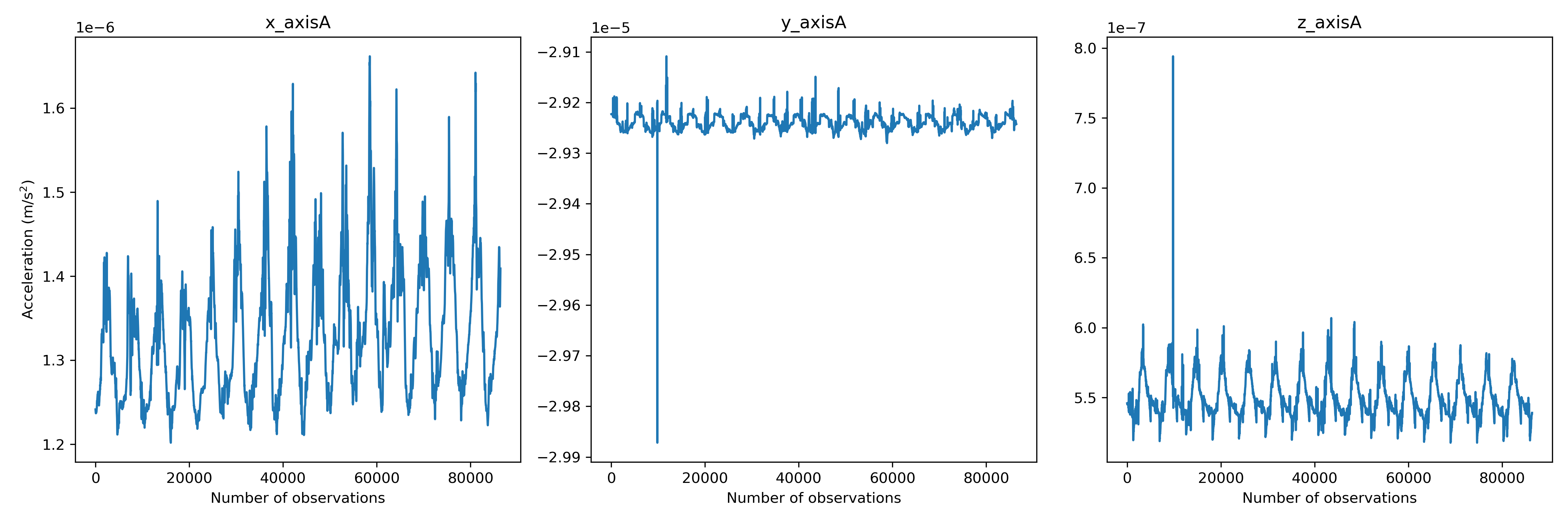}
\includegraphics[scale=.48]{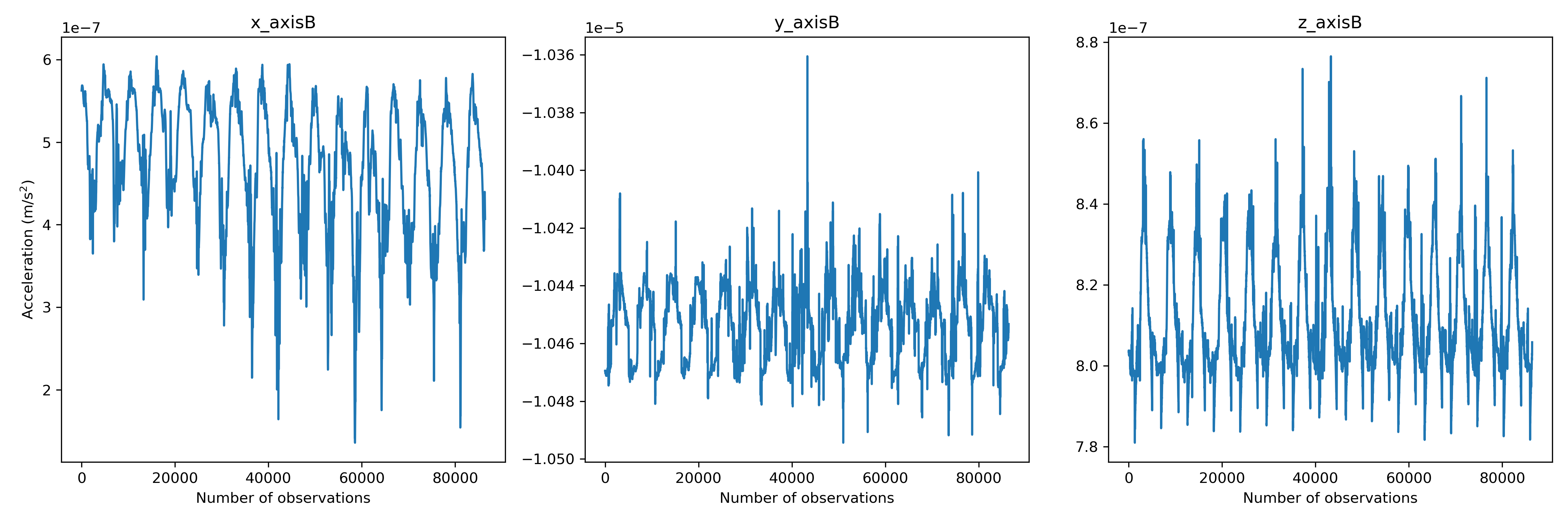}
\caption{Accelerations for 30 May 2005 for GRACE A and GRACE B.}
\label{fig:acc_real}
\end{figure}

Figure \ref{fig:acc_real} displays a representative day of GRACE accelerometer data. Access to GRACE daily accelerometer data is freely available. Similar to other GRACE instrument data, each file contains one day of data, with accelerometer data sampled at one-second intervals, resulting in 86400 samples along each of the three axes. For the convenience of the machine learning community, the GRACE data used in this paper has been downloaded and made accessible on Kaggle (\url{https://www.kaggle.com/datasets/nedadarbeheshti/grace-satellites-asc-files}).

\subsection{Data Preparation}
As shown in Figure \ref{fig:acc_real}, some of these axes exhibit outliers, data points that deviate significantly from the overall pattern. Outliers can distort statistical measures and adversely affect the performance of machine learning models during training. Therefore, as a preprocessing step prior to training, we apply an outlier removal procedure to ensure the integrity of our data.

In this procedure, we first identify the quartiles, specifically the first quartile ($Q1$) and the third quartile ($Q3$), which divide the data into four equal parts. The interquartile range ($IQR$) is then calculated as the difference between $Q3$ and $Q1$. Based on the $IQR$, we determine acceptable upper and lower limits for data points that should not be considered outliers. These limits are obtained by adding and subtracting a multiple (usually 1.5) of the $IQR$ from the quartiles.

\begin{figure}[H]
\center
\includegraphics[scale=.48]{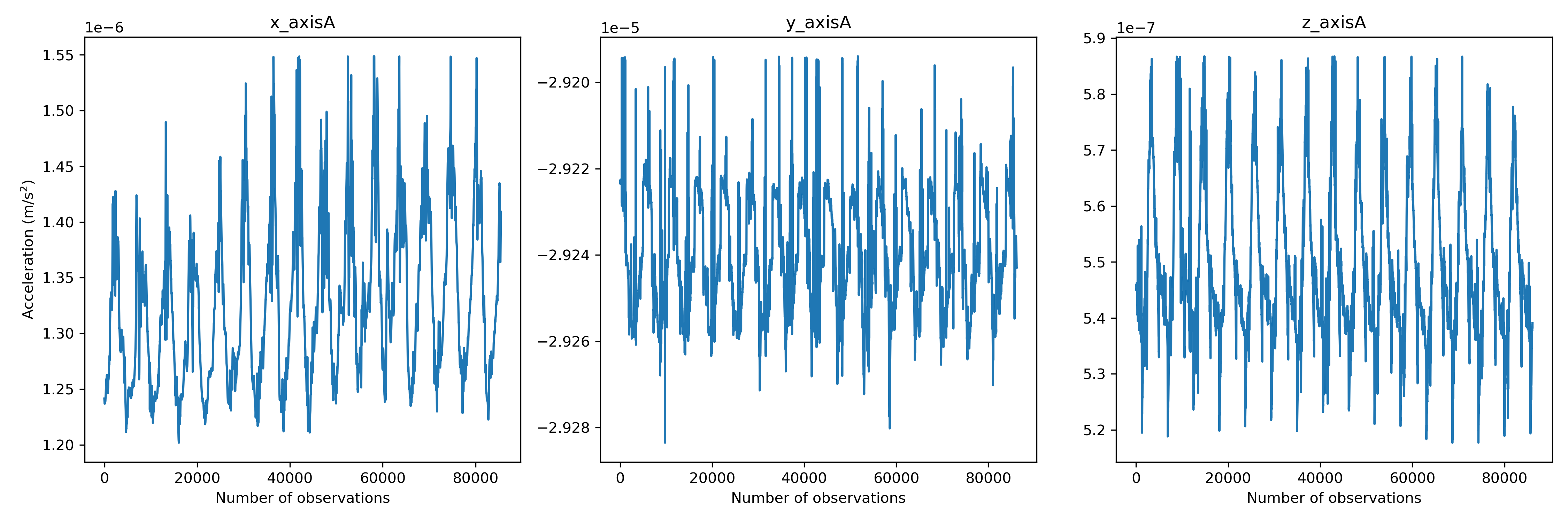}
\includegraphics[scale=.48]{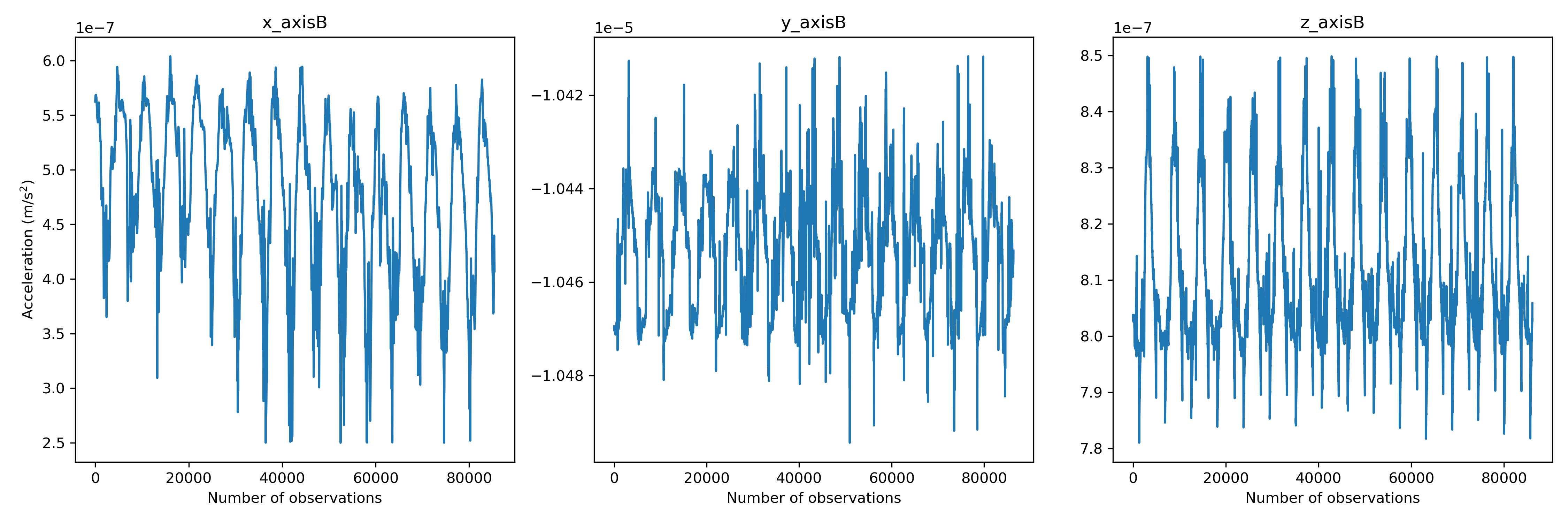}
\caption{Accelerations after outliers removed for 30 May 2005 for GRACE A and GRACE B.}
\label{fig:acc_clean}
\end{figure}

Any data points that fall beyond these limits are considered outliers and are subsequently removed from the dataset. This process helps us mitigate the impact of extreme values on our analysis and ensures that our subsequent machine learning models are more robust and accurate. The function $remove\_outliers(axis)$ implements this outlier removal process. 
\begin{verbatim}
def remove_outliers(axis):
    q1 = np.percentile(axis, 25)
    q3 = np.percentile(axis, 75)
    IQR = q3 - q1
    max_limit = q3 + (1.5 * IQR)
    min_limit = q1 - (1.5 * IQR)
    outliers = np.logical_or(axis > max_limit, axis < min_limit)
    cleaned_data = axis[~outliers]
    return cleaned_data
\end{verbatim}
Given a time series along a specific axis, the function calculates the quartiles, determines the acceptable limits, and filters out any data points that lie beyond these limits. The resulting 'cleaned' time series, shown in the Figure \ref{fig:acc_clean}, allows us to train our models on more reliable and meaningful data, ultimately leading to improved results during the analysis phase.

The next step is to normalize the data. We normalize each time series individually using the \texttt{sklearn.preprocessing} module. We tested two normalization methods, namely \texttt{MinMaxScaler} and  \texttt{RobustScaler}.

\texttt{MinMaxScaler:} This normalization technique transforms each feature individually by scaling it to a given range, usually between 0 and 1. It works by subtracting the minimum value of the feature and then dividing by the range of the feature values. This method is sensitive to outliers, as extreme values can affect the scaling of the entire feature. MinMaxScaler is particularly useful when features have a known range and need to be transformed to a common scale, making them suitable for machine learning algorithms that rely on feature scaling.

\texttt{RobustScaler:} In contrast to MinMaxScaler, RobustScaler is robust to outliers in the data. It scales features using statistics that are robust to outliers, specifically the median and the interquartile range (IQR). The median is subtracted from each data point, and the result is divided by the IQR, ensuring that the scaling is less affected by extreme values. RobustScaler is a good choice when dealing with datasets that contain outliers, as it provides a more stable scaling that is less influenced by these anomalies.

In our study, we compared the performance of both normalization techniques, MinMaxScaler and RobustScaler, on the GRACE accelerometer data. Through experimentation, we found that MinMaxScaler produced better results for our training process, likely due to the characteristics of our dataset and the nature of the accelerometer measurements. As a result, our subsequent analysis and model training are based on data normalized using MinMaxScaler. 

The final step of data preparation involves downsampling the data. Our motivation for downsampling is to expedite the training process, considering that for other instruments such as KBR, the data sampling rate is 5 seconds. Data sampling of 10 seconds aligns with what we use for all data inputs to gravity field recovery. Therefore, the sampling rate of the accelerometer data along each axis entering the training is 10 seconds, resulting in approximately 8640 samples along each axis.

\subsection{Data Splitting}
The data splitting process involves dividing the daily time series into training and testing subsets. For this purpose, we use a train-test split ratio of $70\%$, meaning only the first $70\%$ of the daily time series is utilized for training our models. This partitioning strategy helps ensure that the models are trained on a sufficient portion of the data while reserving a subset for unbiased evaluation.

To facilitate this data splitting process, we employ the function \texttt{create\_dataset}. This function is responsible for generating the input-output pairs (features and labels) required for training and testing. Specifically, it takes the daily time series data as input and constructs the appropriate sequences of past observations as inputs (\texttt{X}) and the subsequent observation as the corresponding output (\texttt{Y}). This sequential arrangement allows the models to learn temporal dependencies and patterns present in the accelerometer data.

As a result of this process, we obtain four datasets: \texttt{trainX} and \texttt{trainY} for training, and \texttt{testX} and \texttt{testY} for testing. The training datasets, \texttt{trainX} and \texttt{trainY}, are utilized to optimize the model's parameters during the training phase. The testing datasets, \texttt{testX} and \texttt{testY}, remain unseen by the model during training and are employed to assess the model's generalization performance on unseen data.

By adhering to this data splitting approach, we ensure that our models are effectively trained and evaluated, and we are able to draw meaningful conclusions about their performance in forecasting accelerometer data.

\subsection{LSTM model}

Long Short-Term Memory (LSTM) networks are a type of recurrent neural network (RNN) architecture that excel in capturing temporal dependencies and patterns in sequential data. In this section, we present the configuration and training of our LSTM model for forecasting GRACE accelerometer data.

The LSTM model is defined using the Keras framework. The model architecture consists of several layers that collectively learn and represent complex relationships within the input data. Here is a breakdown of the model structure:

\begin{itemize}
    \item A look-back window of size 15 is used to consider the previous 15 time steps as input features.
    \item The model begins with an LSTM layer with 8 units. The input shape is set to (1, look\_back), reflecting the univariate nature of the time series.
    \item Following the LSTM layer, the model includes three dense layers. The first dense layer has 10 units, the second has 32 units, and the final output layer has 1 unit, which predicts the next value in the time series.
    \item The model is compiled using the Adam optimizer with a learning rate of 0.001. The mean squared error (MSE) is chosen as the loss function, and the mean absolute error (MAE) is used as a metric to evaluate the model's performance during training.
    \item The training data (\texttt{trainX} and \texttt{trainY}) is used to train the model for 300 epochs. The training data is divided into batches of size 8 for efficient optimization. A validation split of $15\%$ is applied, where a portion of the training data is used for validation during each epoch.
\end{itemize}

The provided Python code snippet illustrates the implementation of the LSTM model:

\begin{verbatim}
look_back = 15
model = Sequential()
model.add(LSTM(8, input_shape=(1, look_back)))
model.add(Dense(10))
model.add(Dense(32))
model.add(Dense(1))
model.compile(optimizer=Adam(learning_rate=0.001), loss='mse', metrics=["mae"])
history = model.fit(trainX, trainY, epochs=300, batch_size=8, verbose=0, 
                    validation_split=0.15)
\end{verbatim}

\begin{figure}[H]
\center
\includegraphics[scale=.55]{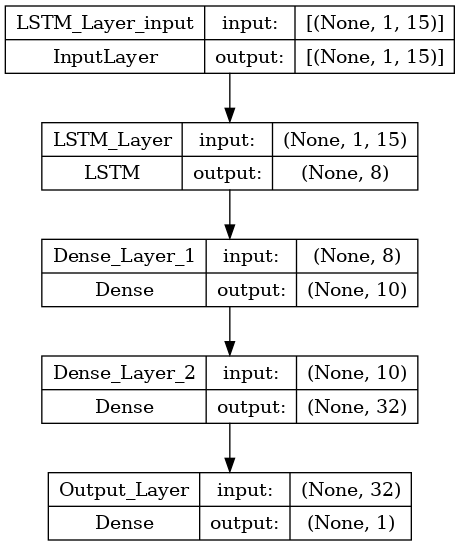}
\caption{Diagram of the LSTM network.}
\label{fig:lstm}
\end{figure}

Throughout the training process, the model learns to capture the temporal relationships and dependencies present in the accelerometer data. The validation results monitored by the MAE metric provide insights into the model's ability to generalize to unseen data. The history object (\texttt{history}) stores information about the training progress, allowing for further analysis and visualization of performance trends. Figure \ref{fig:lstm} presents the LSTM network.

\section{Results}

\begin{figure}[H]
\center
\includegraphics[scale=.35]{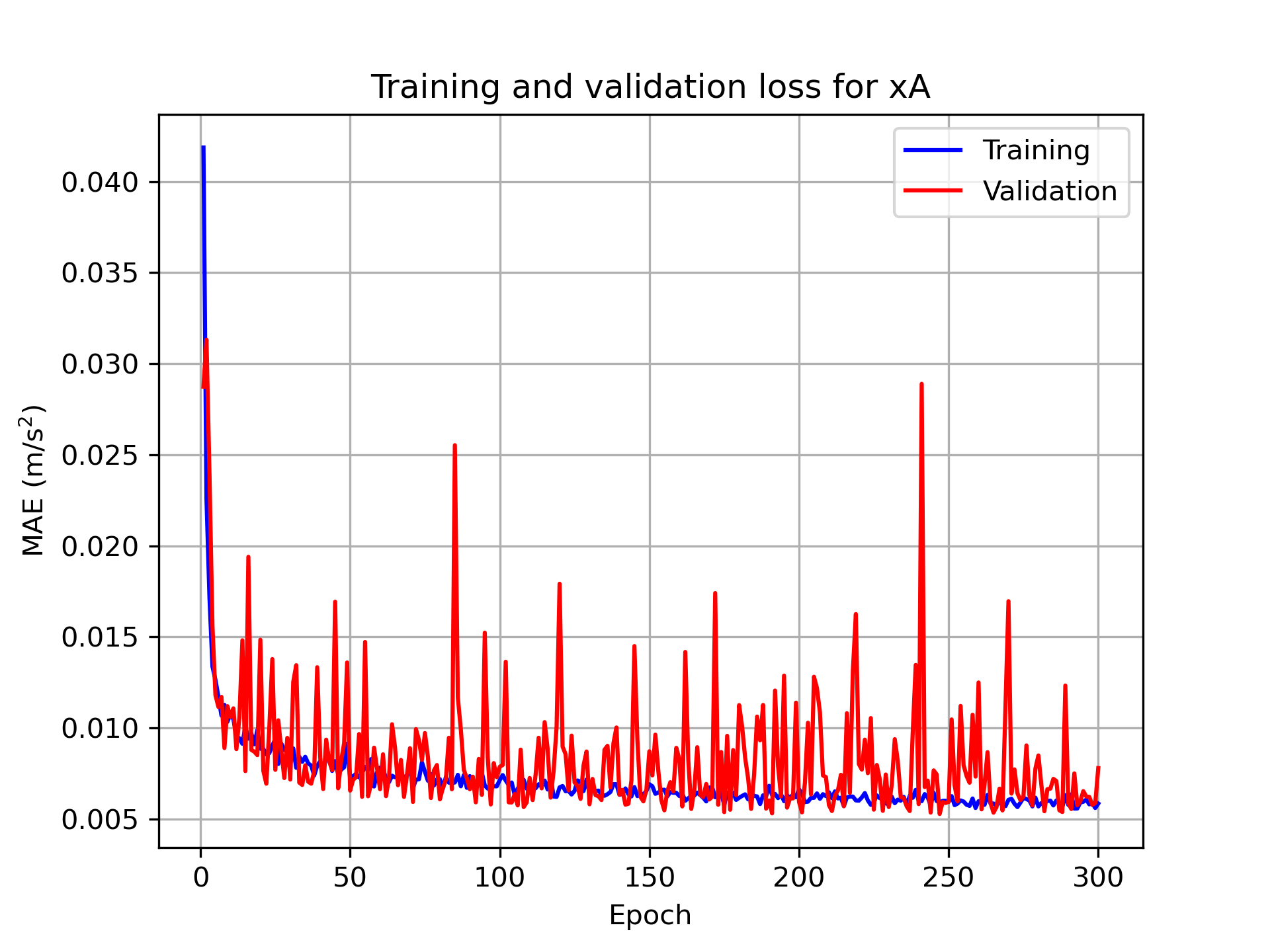}
\includegraphics[scale=.35]{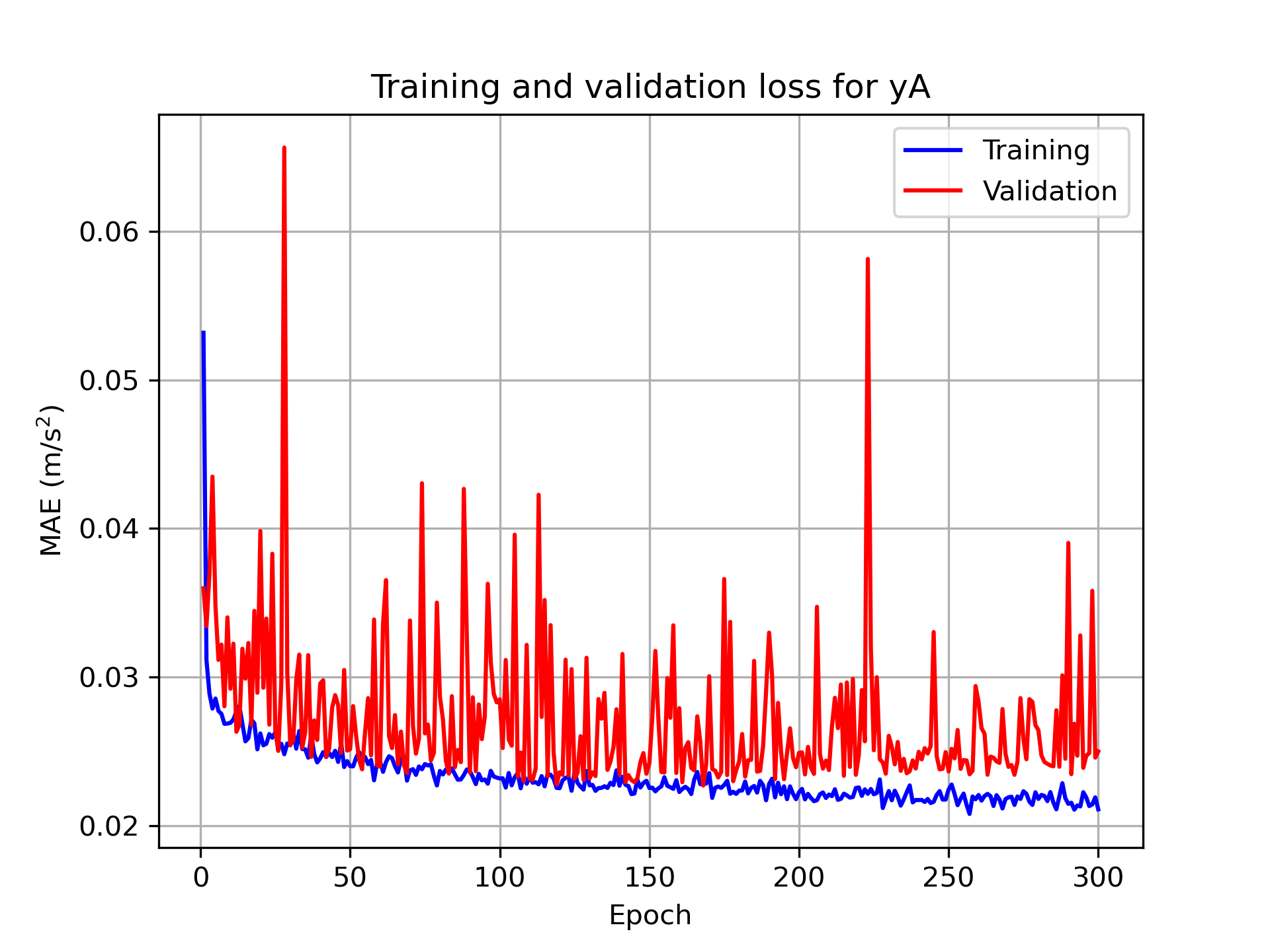}
\includegraphics[scale=.35]{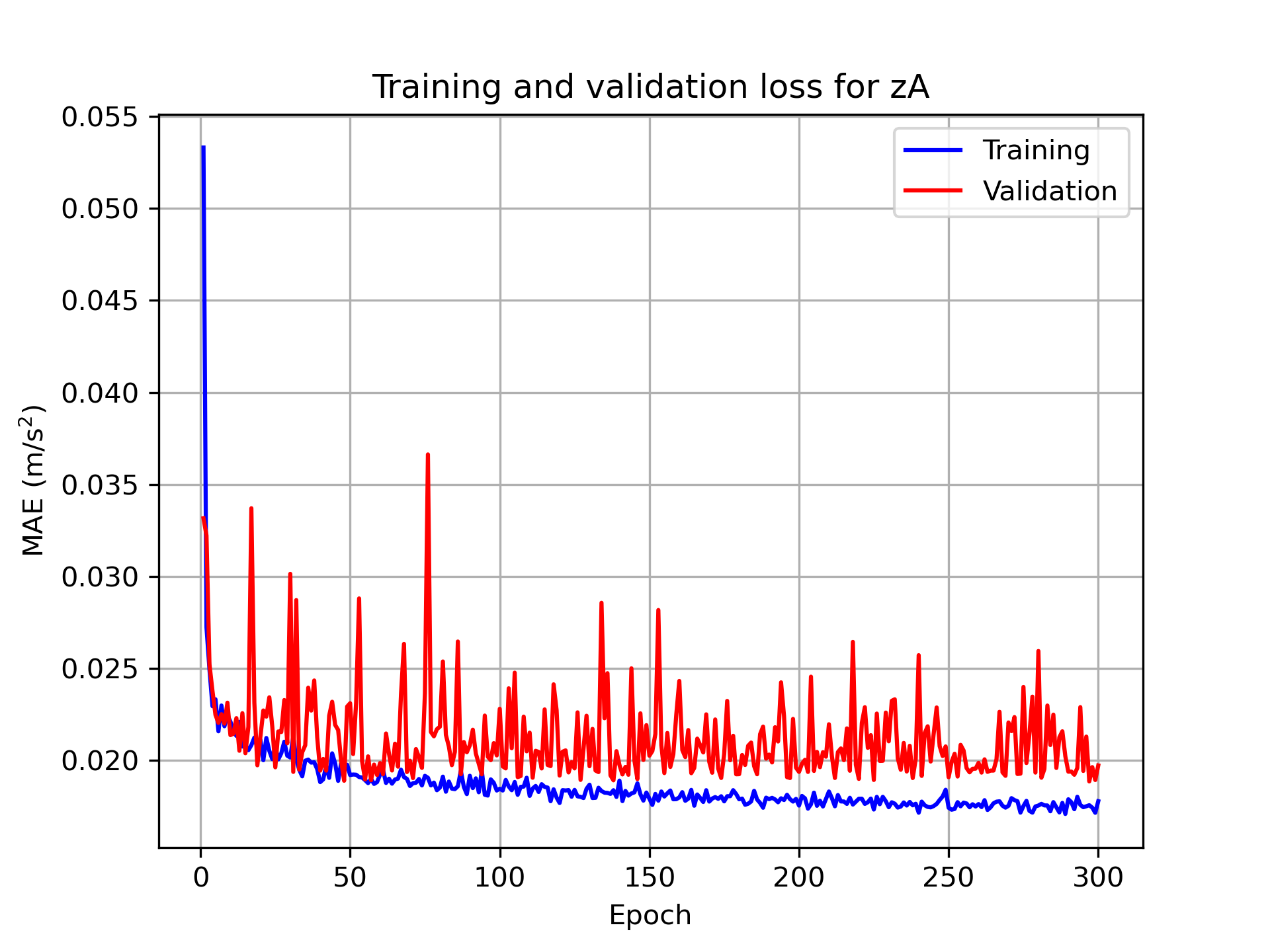}
\includegraphics[scale=.35]{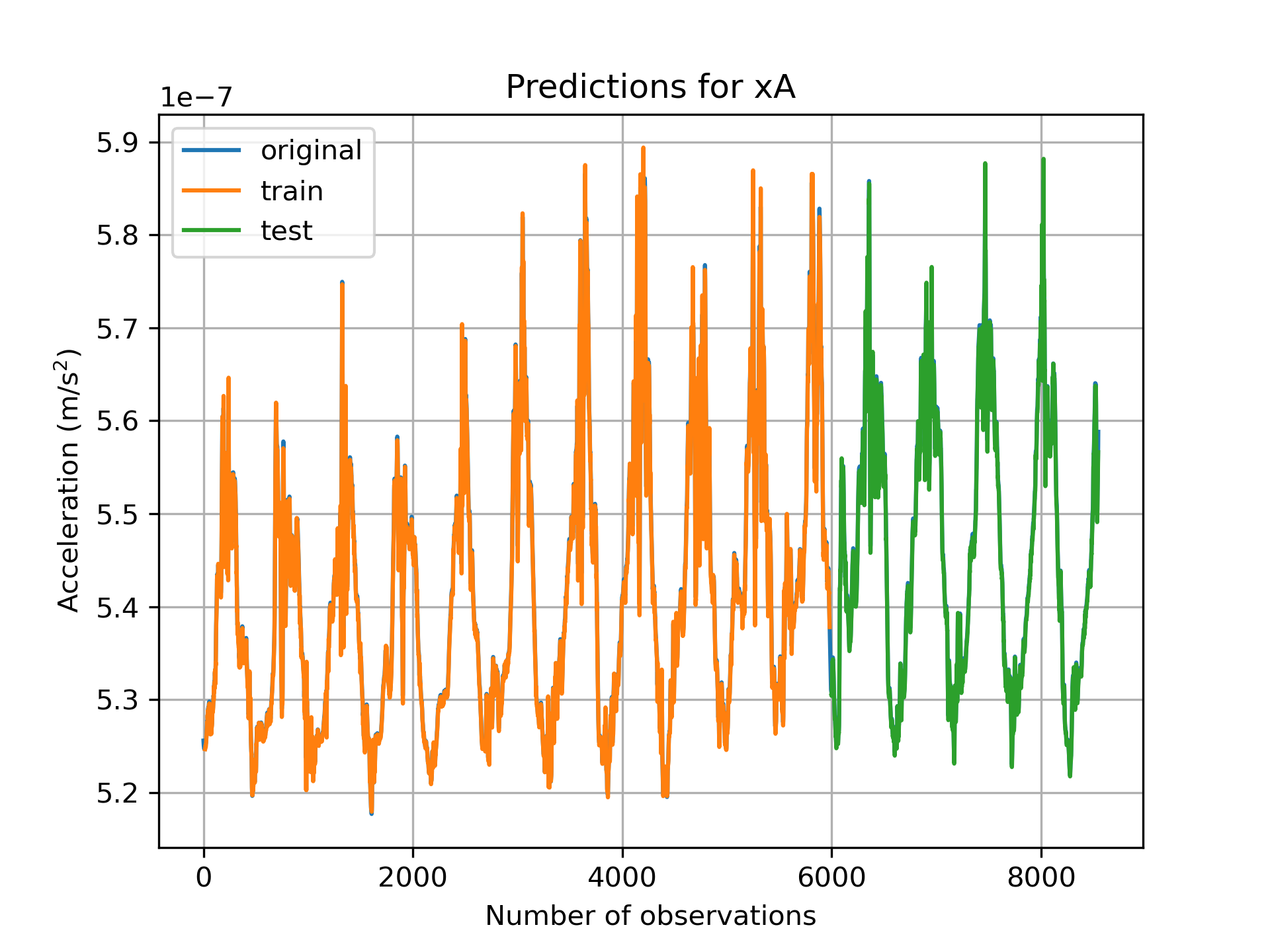}
\includegraphics[scale=.35]{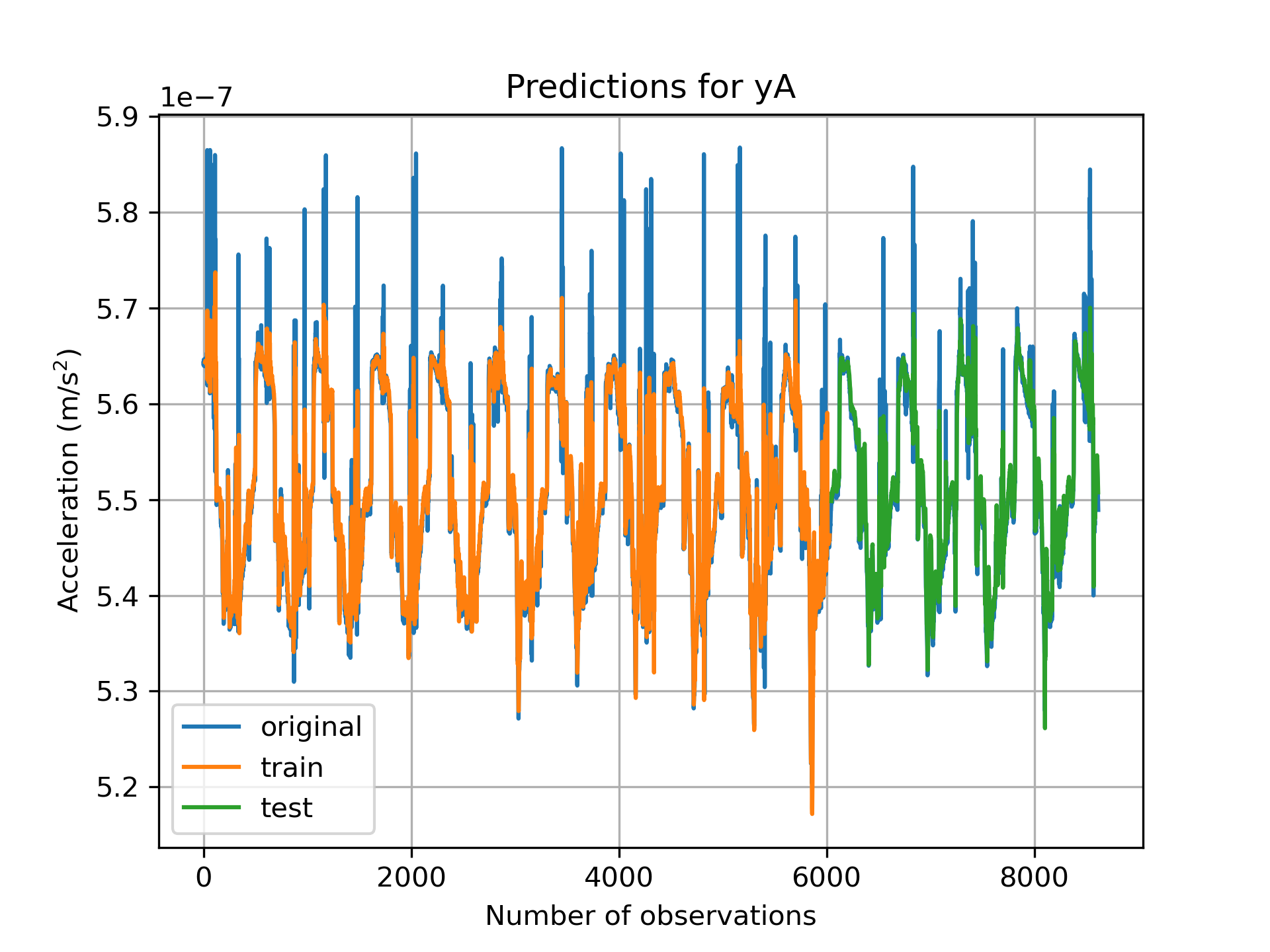}
\includegraphics[scale=.35]{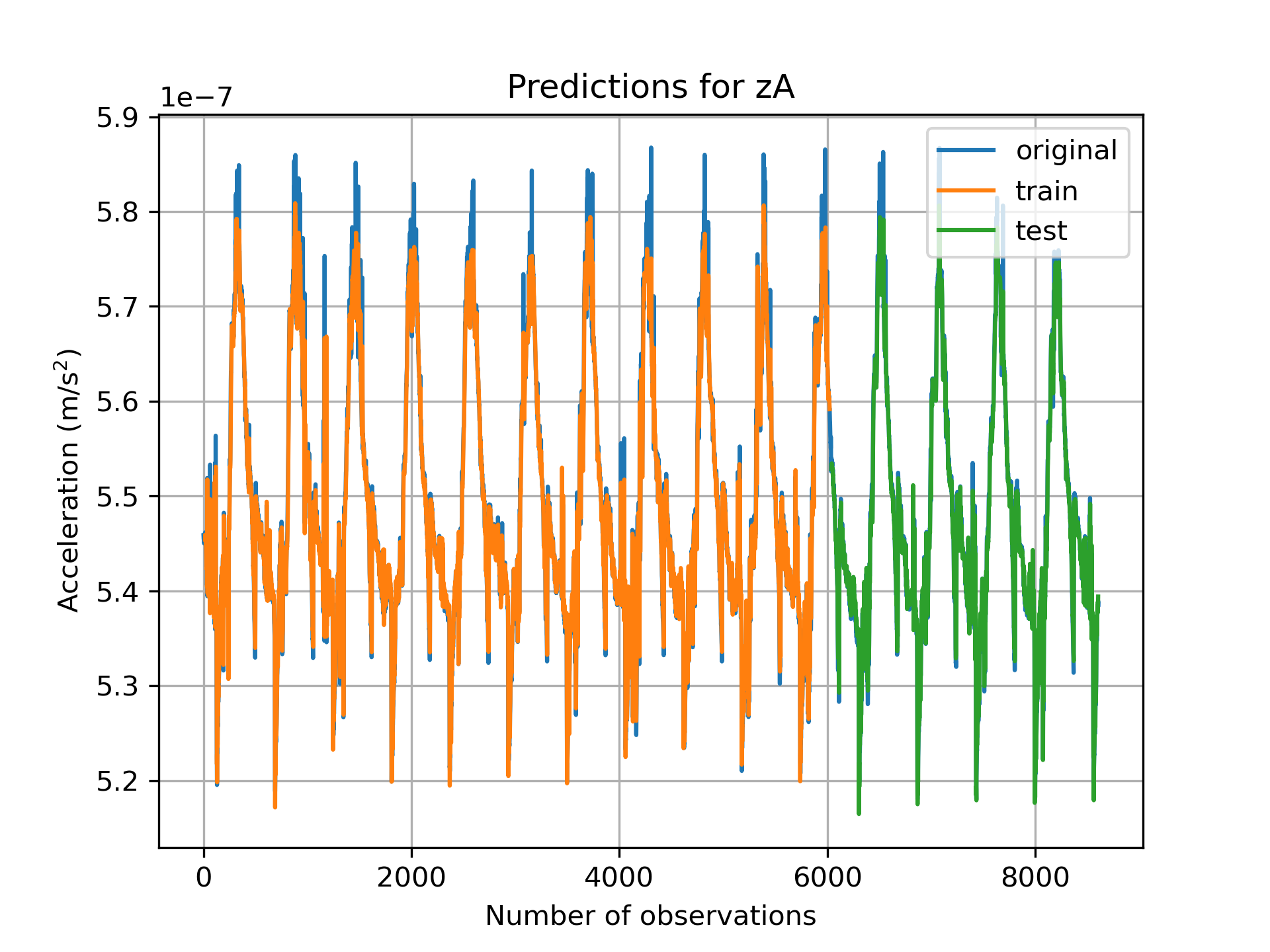}
\includegraphics[scale=.35]{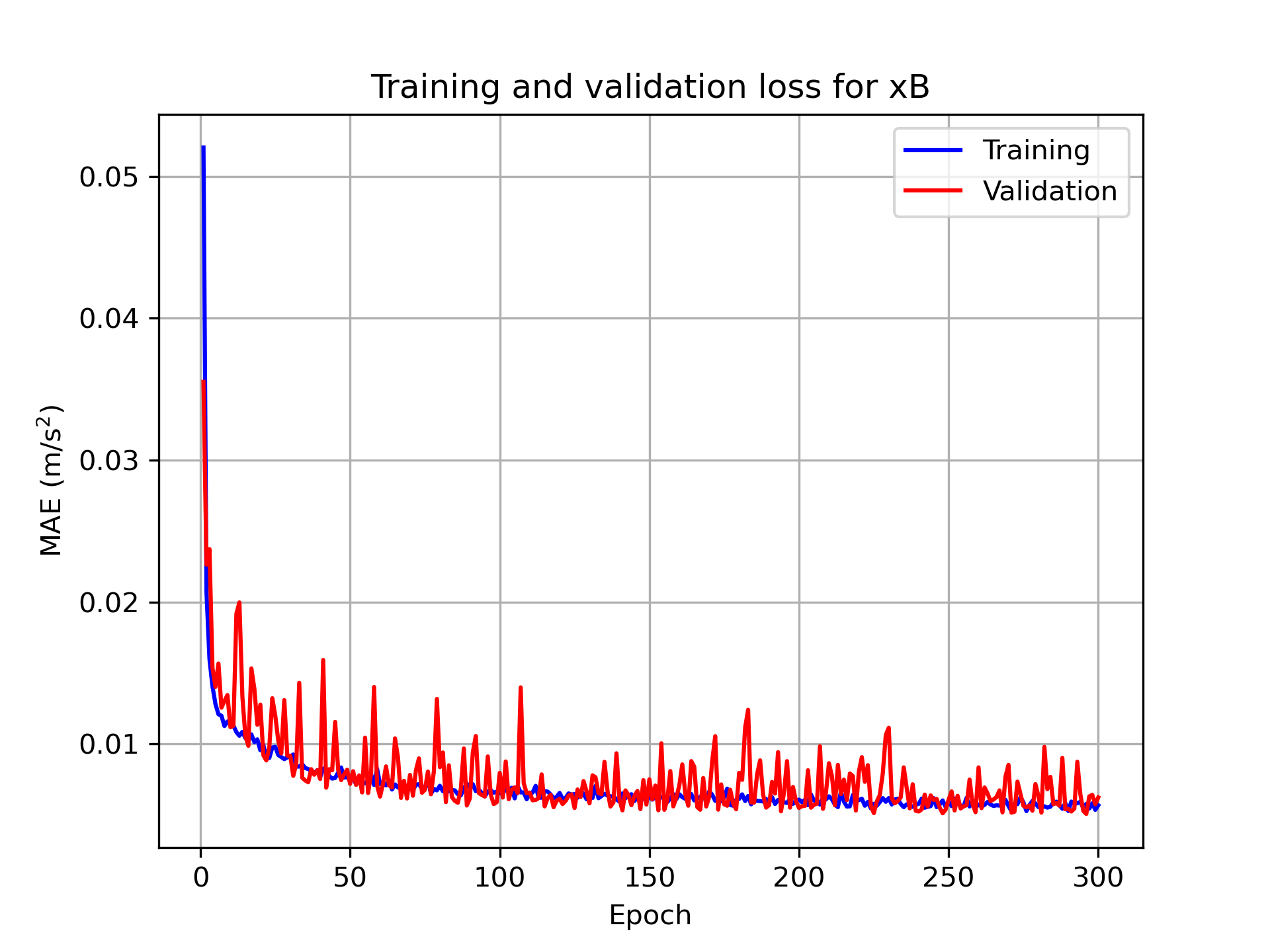}
\includegraphics[scale=.35]{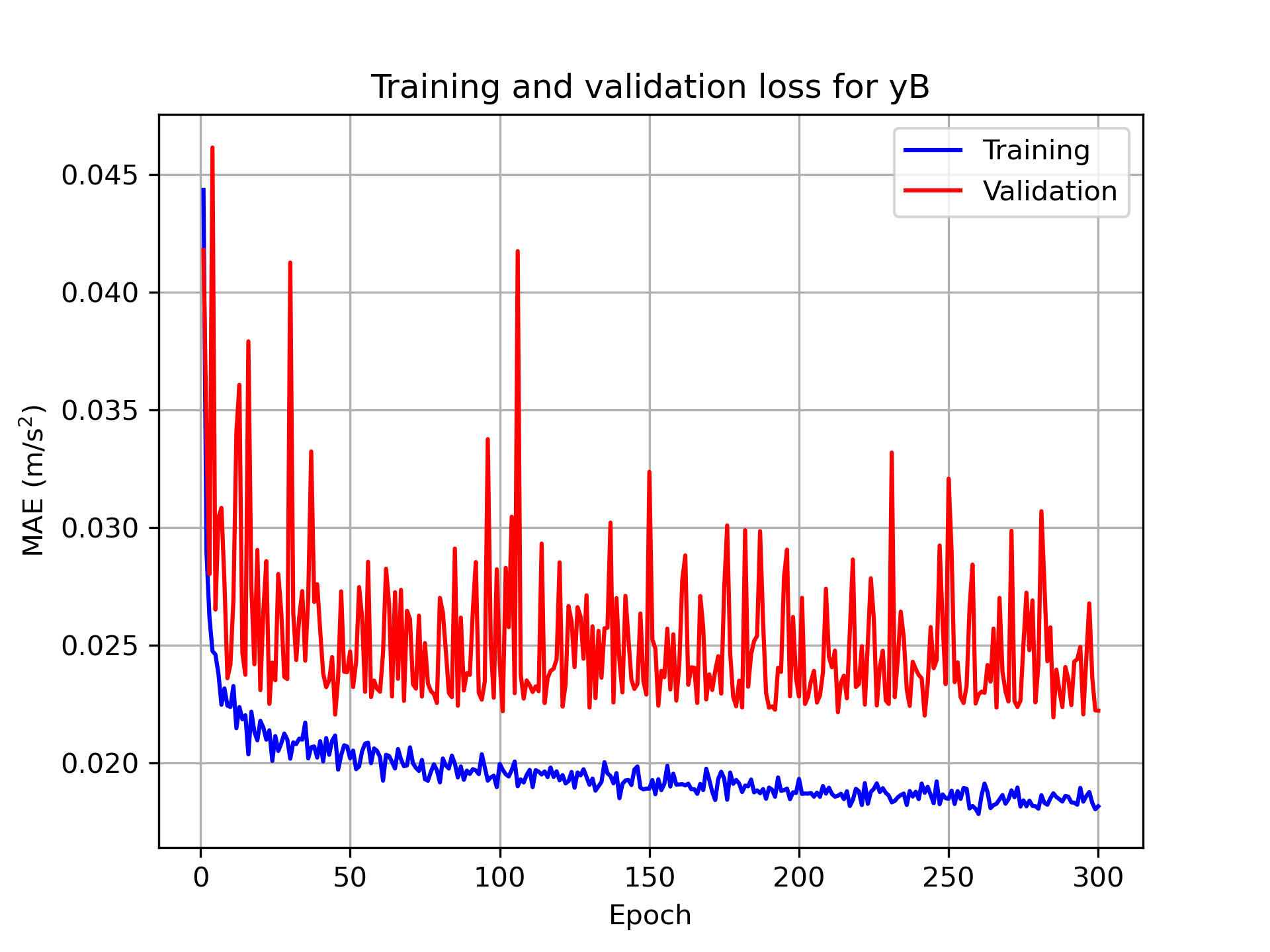}
\includegraphics[scale=.35]{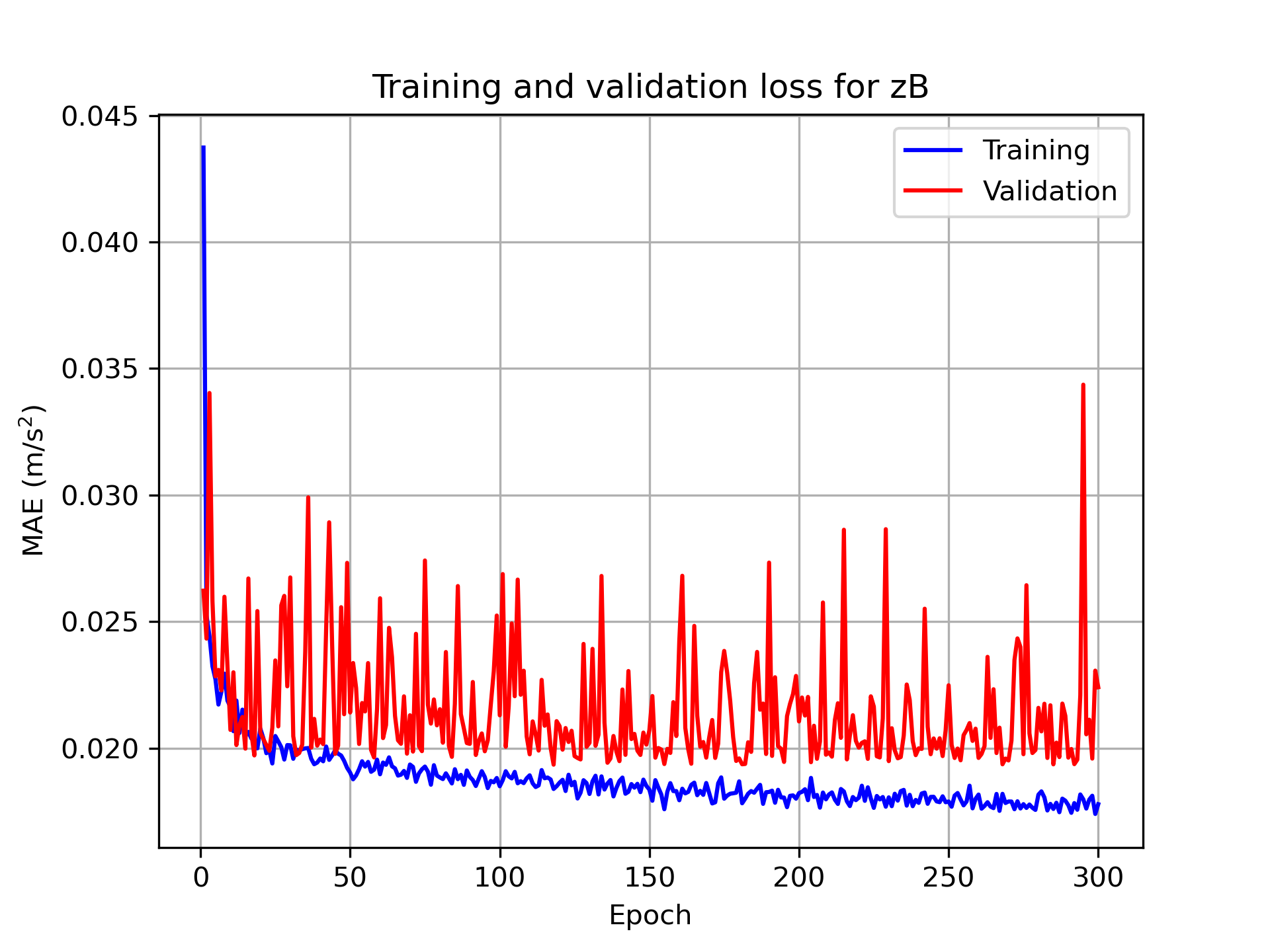}
\includegraphics[scale=.35]{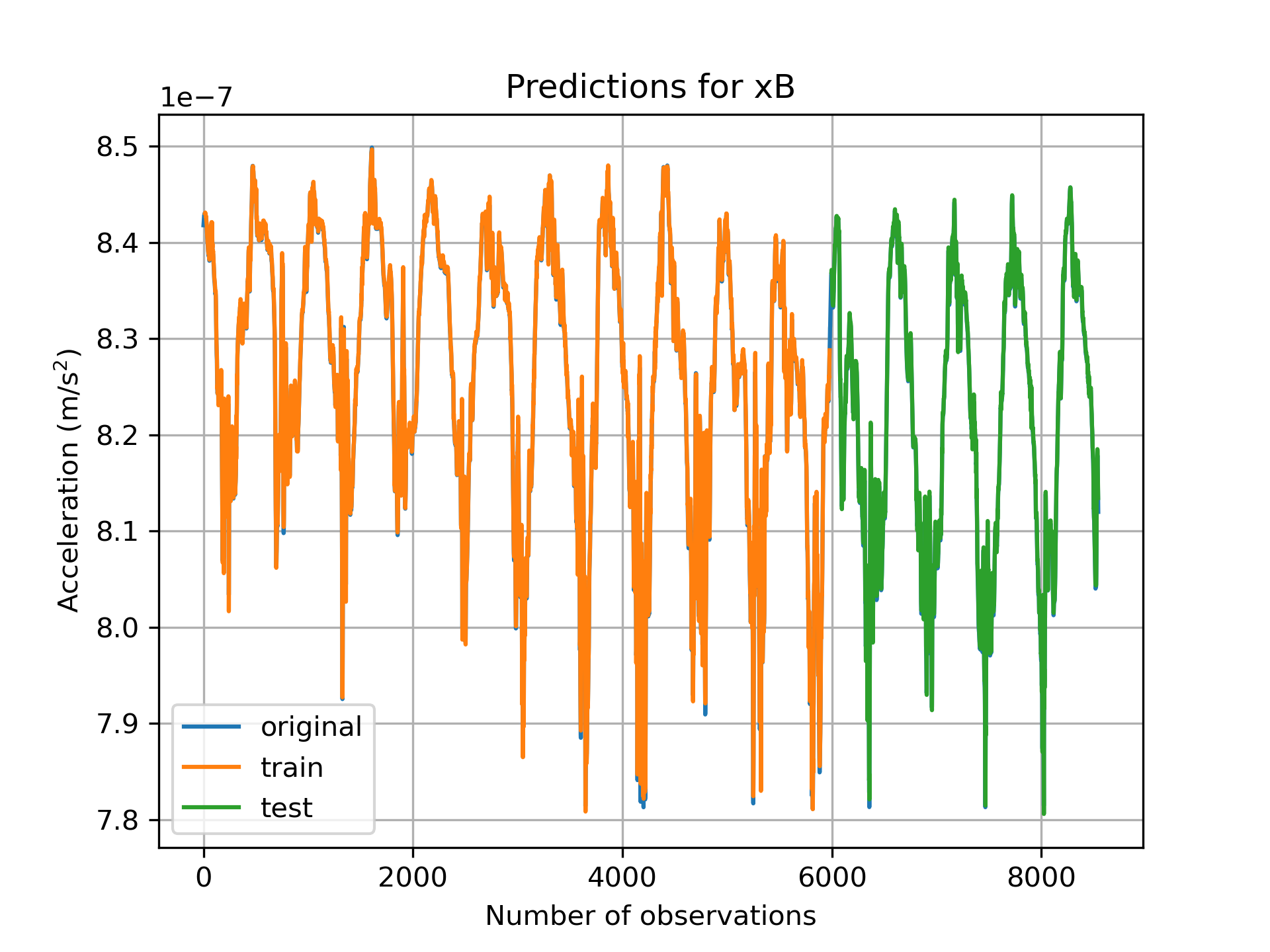}
\includegraphics[scale=.35]{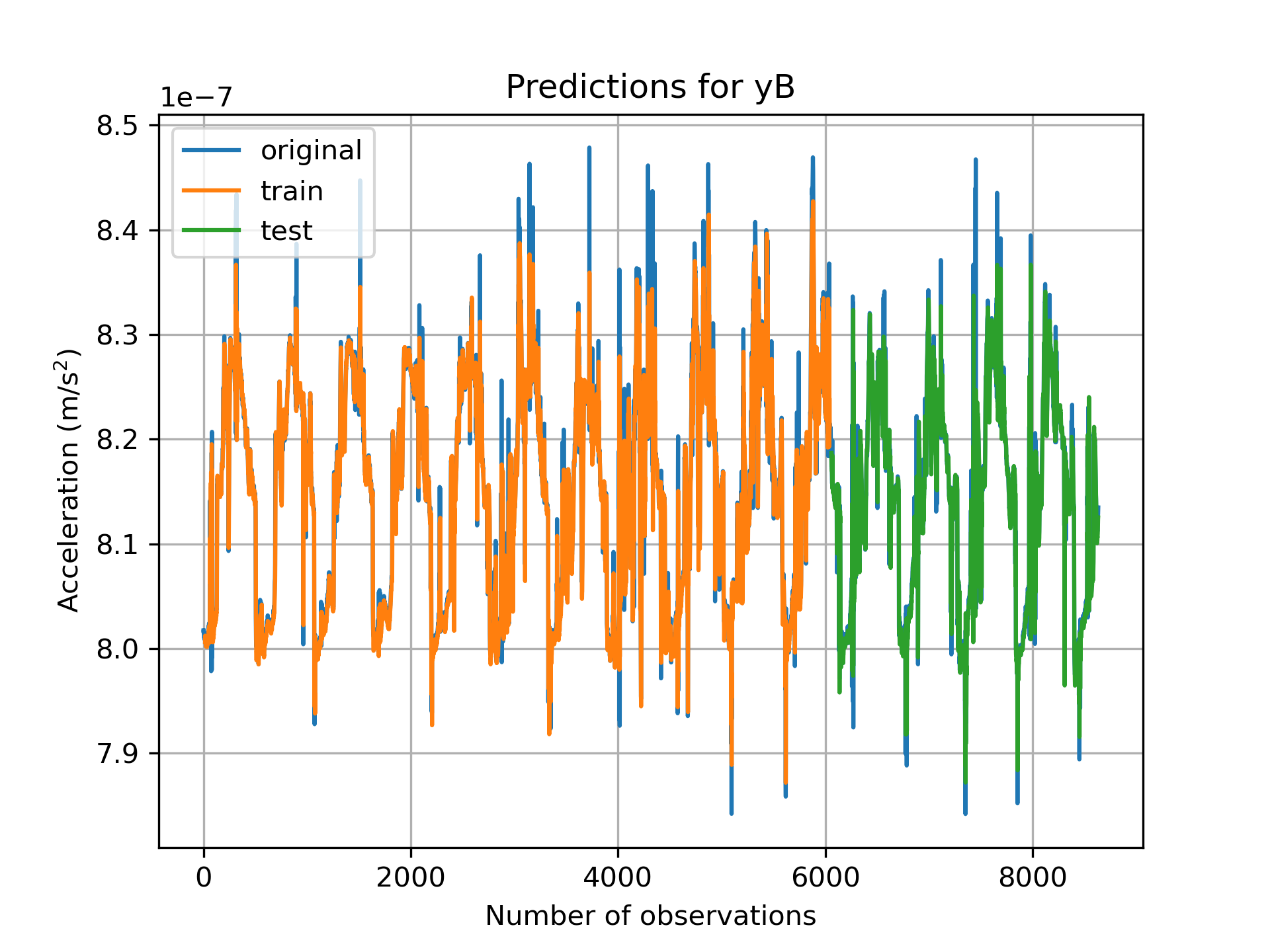}
\includegraphics[scale=.35]{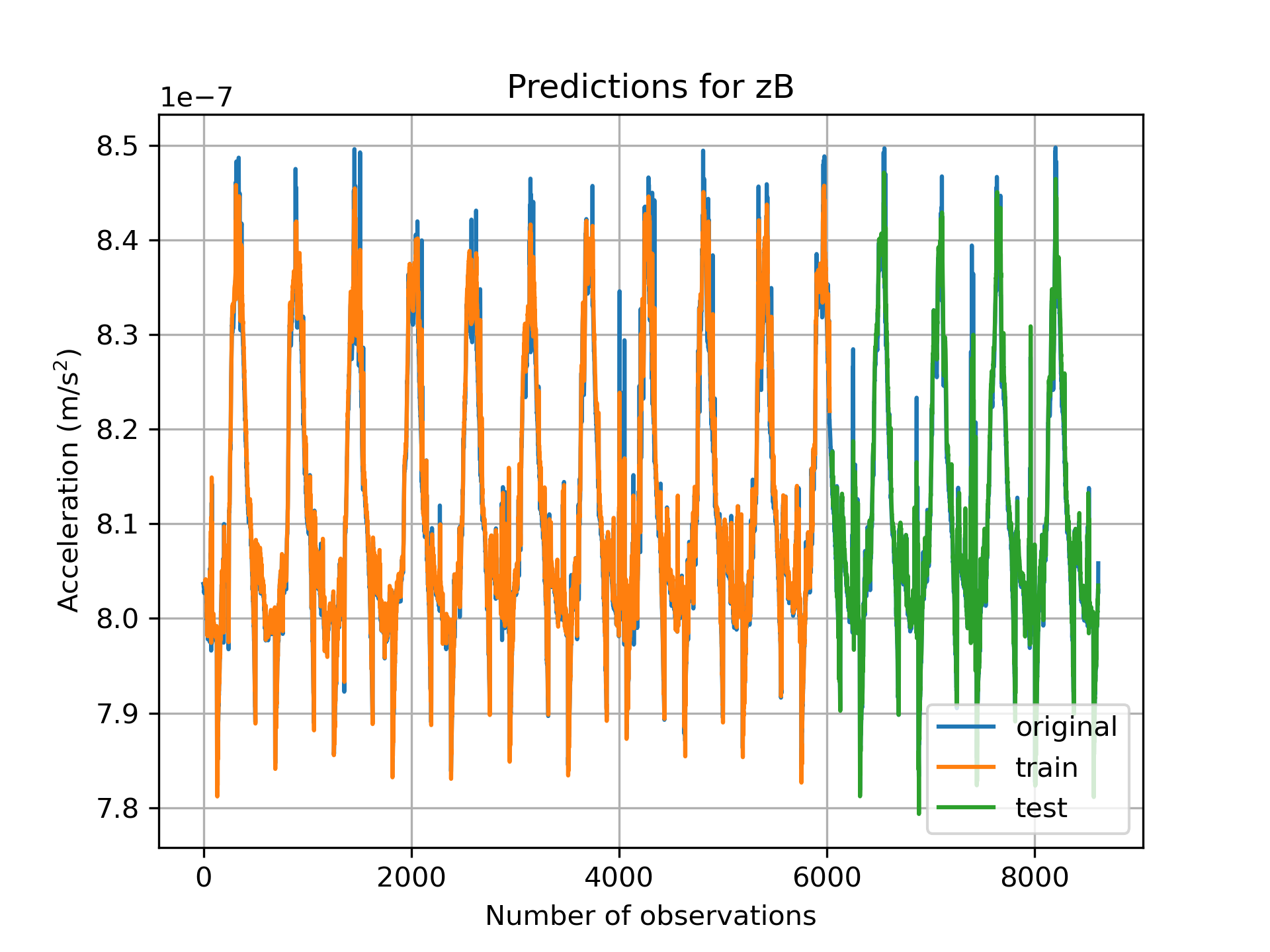}
\caption{LSTM loss and prediction plots for each axis per satellite.}
\label{fig:loss}
\end{figure}

Figure \ref{fig:loss} illustrates the loss and prediction plots for each axis of the accelerometer data. Across all axes, the training loss consistently remains lower than the validation loss. This observed trend underscores the efficacy of the employed outlier removal procedure, demonstrating successful mitigation of overfitting. Notably, the loss plot demonstrates a more favorable convergence for the $x$ axis, evident in both satellites. This convergence occurs at a level below 0.01 MAE for both training and validation datasets.

\begin{figure}[H]
\center
\includegraphics[scale=.35]{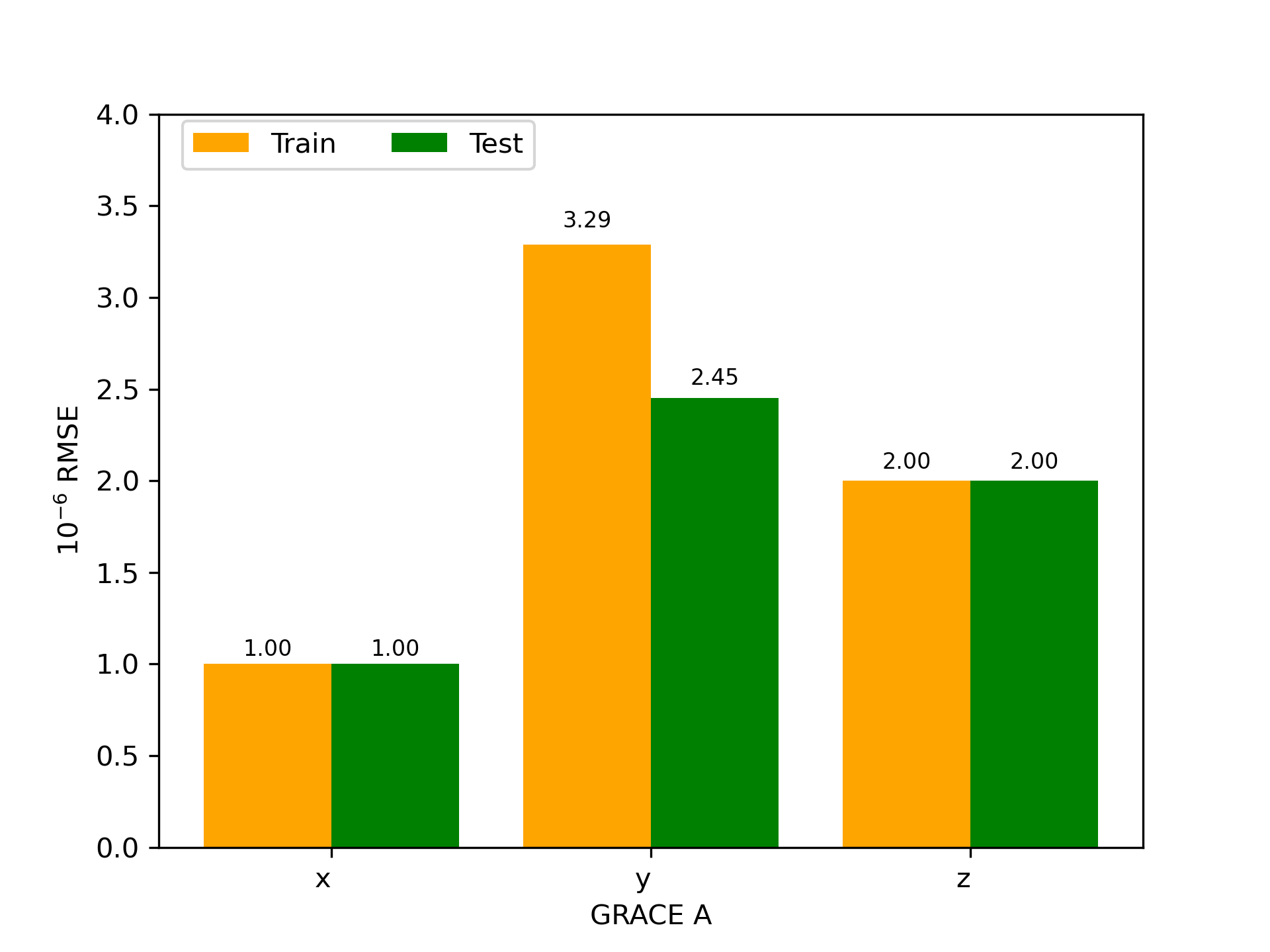}
\includegraphics[scale=.35]{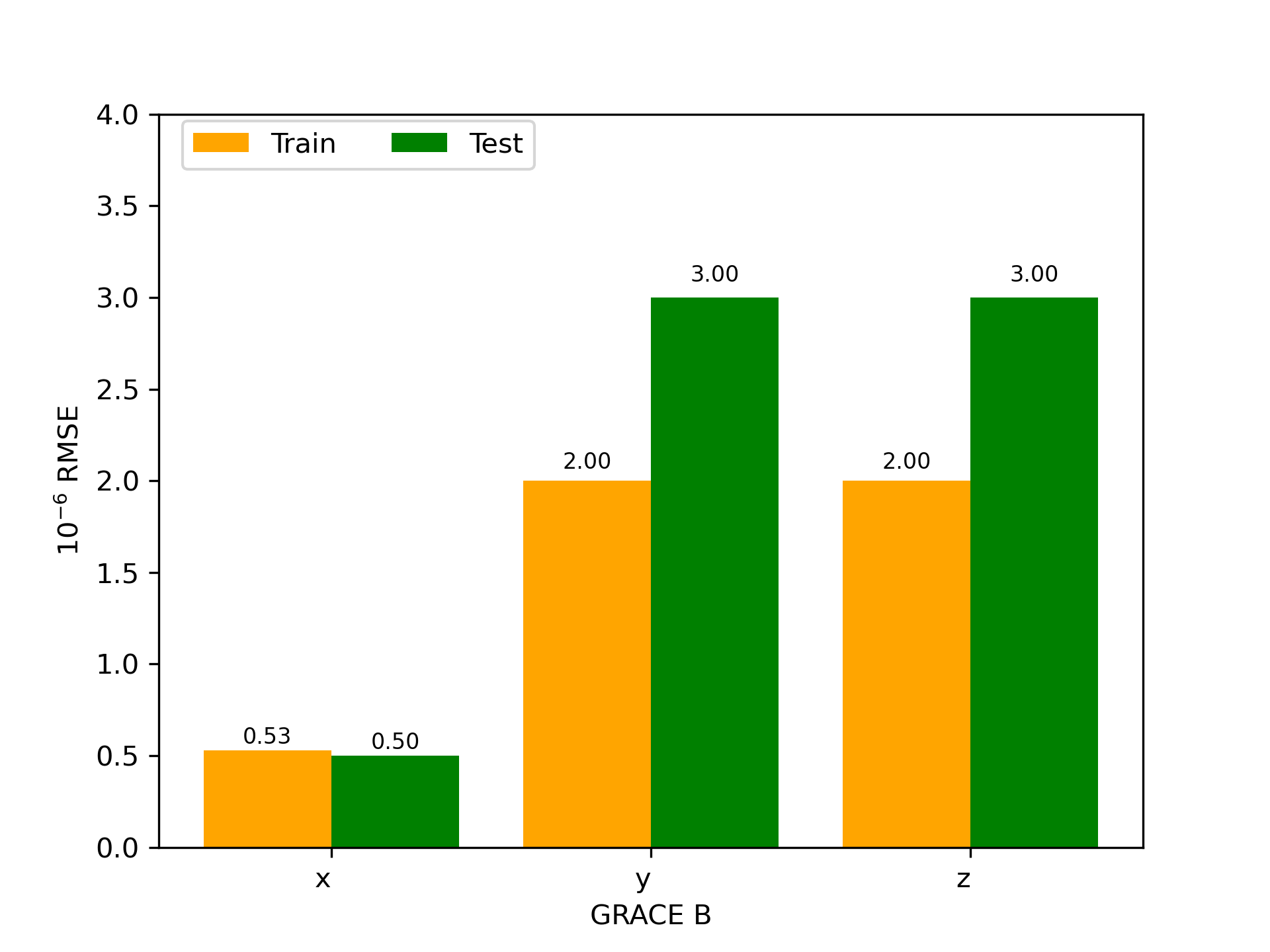}
\caption{Bar plot of RMSE scores for each axis.}
\label{fig:bar}
\end{figure}

It is worth highlighting that for the $y$ and $z$ axes of GRACE B, discernible disparities emerge between the training and validation loss. This distinction is further underscored by the divergent RMSE scores, as indicated in Table \ref{tab:rmse-scores} and visually depicted in Figure \ref{fig:bar}. The presence of differences between the RMSE scores for training and testing reinforces the noteworthy variability observed in the y and z axes of GRACE B.

Table \ref{tab:rmse-scores} furnishes additional details concerning the total number of data points subsequent to the outlier removal process. It is noteworthy that the root mean square error (RMSE) scores for both training and testing phases exhibit a remarkable degree of uniformity across each axis. This uniformity signifies that the executed outlier removal procedure adeptly preserves equilibrium between the training and test subsets, further affirming the procedure's efficacy in facilitating a balanced distribution of data points.

\begin{table}[h]
\centering
\caption{The total number of data points post outlier removal and RMSE scores for each axis.}
\begin{tabular}{|c|c|c|c|}
\hline
\multicolumn{4}{|c|}{\textbf{GRACE A}} \\
\hline
\textbf{Axis} & \textbf{Size} & \textbf{Train Score ($10^{-6}$ RMSE)} & \textbf{Test Score ($10^{-6}$ RMSE)} \\
\hline
$x$ & (8548) & $1$ & $1$ \\
$y$ & (8618) & $3.29$ & $2.45$ \\
$z$ & (8608) & $2$ & $2$ \\
\hline
\end{tabular}
\quad
\begin{tabular}{|c|c|c|c|}
\hline
\multicolumn{4}{|c|}{\textbf{GRACE B}} \\
\hline
\textbf{Axis} & \textbf{Size} & \textbf{Train Score ($10^{-6}$ RMSE)} & \textbf{Test Score ($10^{-6}$ RMSE)} \\
\hline
$x$ & (8545) & $0.53$ & $0.50$ \\
$y$ & (8634) & $2$ & $3$ \\
$z$ & (8613) & $2$ & $3$ \\
\hline
\end{tabular}
\label{tab:rmse-scores}
\end{table}

The comparable performance between training and testing scores across axes further affirms the robustness of our approach. It indicates that the LSTM model has learned the underlying patterns and dependencies in the accelerometer data without excessively memorizing the noise or anomalies.

\section{Conclusion and Future Directions}
In the context of this paper, we have embarked on an the application of Long Short-Term Memory (LSTM) networks for forecasting GRACE accelerometer data, a concept with implications for the broader realm of GRACE instruments. The success of this endeavor lays the groundwork for extending this approach to other key instruments within the GRACE mission, including GPS, KBR (K-Band Ranging), and  star cameras.

Moving forward, our next phase involves implementing this approach to create time series for each instrument over a month's span. This study aims to assess the feasibility of utilizing LSTM networks in terms of computational resources and time efficiency to forecast data gaps across various instruments.

This study is a significant step toward the development of a specialized Generative Pre-trained Transformer (GPT) tailored specifically for the GRACE type missions. This holds the potential to generate accurate time series data for each instrument within defined temporal boundaries. 

\codedataavailability{Data and code are available at
\url{https://www.kaggle.com/datasets/nedadarbeheshti/grace-satellites-asc-files} and
\url{https://github.com/Darbeheshti/LSTM-based-Analysis-for-GRACE-Accelerometers} }

 
\noappendix       


\authorcontribution{ 
Elahe Moradi reviewed the LSTM code and the article.
Neda Darbeheshti performed the GRACE data processing and developed LSTM code and prepared the manuscript with contributions from Elahe Moradi.} 

\competinginterests{The authors declare that they have no conflict of interest.} 




\bibliographystyle{copernicus}
\bibliography{bib4aiubrl03grace.bib}

\begin{thebibliography}{14}
\providecommand{\natexlab}[1]{#1}
\providecommand{\url}[1]{{\tt #1}}
\providecommand{\urlprefix}{URL }
\expandafter\ifx\csname urlstyle\endcsname\relax
  \providecommand{\doi}[1]{doi:\discretionary{}{}{}#1}\else
  \providecommand{\doi}{doi:\discretionary{}{}{}\begingroup
  \urlstyle{rm}\Url}\fi

\bibitem[{Ahi and Cekim(2021)}]{Ahi2021}
Ahi, G. and Cekim, H.: {Long-term temporal prediction of terrestrial water
  storage changes over global basins using GRACE and limited GRACE-FO data},
  Acta Geodaetica et Geophysica, pp. 321--344,
  \doi{10.1007/s40328-021-00338-4}, 2021.

\bibitem[{Bandikova et~al.(2019)Bandikova, McCullough, Kruizinga, Save, and
  Christophe}]{Bandikova2019}
Bandikova, T., McCullough, C., Kruizinga, G.~L., Save, H., and Christophe, B.:
  GRACE accelerometer data transplant, Advances in Space Research, 64, Issue 3,
  2019.

\bibitem[{Behzadpour et~al.(2021)Behzadpour, , Mayer~G{\"u}rr, and
  Krauss}]{Behzadpour2021}
Behzadpour, S., , Mayer~G{\"u}rr, T., and Krauss, S.: GRACE Follow-On
  accelerometer data recovery, Journal of Geophysical Research: Solid Earth,
  126, 2021.

\bibitem[{Chollet(2017)}]{Chollet2017}
Chollet, F.: Deep Learning with Python, Manning Publications, 2017.

\bibitem[{Darbeheshti et~al.(2017)Darbeheshti, Wegener, M{\"u}ller, Naeimi,
  Heinzel, and Hewitson}]{darbeheshti2017}
Darbeheshti, N., Wegener, H., M{\"u}ller, V., Naeimi, M., Heinzel, G., and
  Hewitson, M.: Instrument data simulations for GRACE Follow-on: observation
  and noise models, Earth System Science Data 9, \doi{10.5194/essd-9-833-2017},
  2017.

\bibitem[{Darbeheshti et~al.(2023)Darbeheshti, Lasser, Meyer, Arnold, and
  J{\"a}ggi}]{darbeheshti2023}
Darbeheshti, N., Lasser, M., Meyer, U., Arnold, D., and J{\"a}ggi, A.:
  AIUB-GRACE gravity field solutions for G3P: processing strategies and
  instrument parametrization, Earth Syst. Sci. Data Discuss. preprint,
  \doi{doi.org/10.5194/essd-2023-72}, 2023.

\bibitem[{Gou et~al.(2023)Gou, Kiani~Shahvandi, Hohensinn, and Soja}]{Gou2023}
Gou, J., Kiani~Shahvandi, M., Hohensinn, R., and Soja, B.: Ultra-short-term
  prediction of LOD using LSTM neural networks, Journal of Geodesy, 97,
  \doi{10.1007/s00190-023-01745-x}, 2023.

\bibitem[{Ince et~al.(2019)Ince, Rei{\ss}land, Elger, F{\"o}rste, Flechtner,
  and Schuh}]{Ince2019}
Ince, E. S.and~Barthelmes, F., Rei{\ss}land, S., Elger, K., F{\"o}rste, C.,
  Flechtner, F., and Schuh, H.: ICGEM - 15 years of successful collection and
  distribution of global gravitational models, associated services and future
  plans, Earth System Science Data 11, pp. 647--674,
  \doi{10.5194/essd-11-647-2019}, 2019.

\bibitem[{Kaselimi(2021)}]{Kaselimi2021}
Kaselimi, M.: Machine learning methods for modelling and analysis of time
  series signals in geoinformatics, Ph.D. thesis, National Technical University
  of Athens, 2021.

\bibitem[{Klinger(2018)}]{klinger2018}
Klinger, B.: A contribution to GRACE time-variable gravity field recovery:
  Improved Level-1B data pre-processing methodologies, Ph.D. thesis, Graz
  University of Technology, 2018.

\bibitem[{Klinger and Mayer~G{\"u}rr(2016)}]{klinger2016}
Klinger, B. and Mayer~G{\"u}rr, T.: The role of accelerometer data calibration
  within GRACE gravity field recovery: Results from ITSG-Grace2016, Advances in
  Space Research 58, 2016.

\bibitem[{Tapley et~al.(2019)Tapley, Watkins, Flechtner, and
  et~al.}]{tapley2019}
Tapley, B., Watkins, M., Flechtner, F., and et~al.: Contributions of GRACE to
  understanding climate change., Nat. Clim. Chang., pp. 358--369,
  \urlprefix\url{https://doi.org/10.1038/s41558-019-0456-2}, 2019.

\bibitem[{Tapley et~al.(2004)Tapley, Bettadpur, Watkins, and
  Reigber}]{tapley2004}
Tapley, B.~D., Bettadpur, S., Watkins, M., and Reigber, C.: The gravity
  recovery and climate experiment: Mission overview and early results,
  Geophysical Research Letters, 31, \doi{10.1029/2004GL019920}, l09607, 2004.

\bibitem[{Wahr et~al.(1998)Wahr, Molenaar, and Bryan}]{wahr1998}
Wahr, J., Molenaar, M., and Bryan, F.: Time variability of the Earth's gravity
  field: Hydrological and oceanic effects and their possible detection using
  GRACE, Journal of Geophysical Research: Solid Earth, 103, 30\,205--30\,229,
  \doi{10.1029/98JB02844}, 1998.

\end{thebibliography}

\end{document}